\documentclass[letterpaper, 10pt, conference]{ieeeconf}
\usepackage{times}
\usepackage[pdftex]{graphicx}

\usepackage{amsmath,amssymb,amsopn,amstext,amsfonts, amsthm}
\usepackage{cancel}
\usepackage[space]{cite}
\usepackage{pdfsync}
\usepackage{balance}
\usepackage{color}
\usepackage{mathtools}
\usepackage[ruled,norelsize, linesnumbered]{algorithm2e}
\usepackage{bm}
\usepackage{diagbox}
\usepackage{float}
\usepackage[outdir=./]{epstopdf}
\usepackage{url}
\usepackage{multirow}
\usepackage{makecell}
\usepackage[export]{adjustbox}
\usepackage{commath}
\usepackage{bbold}
\let\labelindent\relax
\usepackage{enumitem}
\usepackage[font=small,skip=2pt]{caption}
\usepackage{subcaption}

\usepackage[linkcolor=black,citecolor=black,urlcolor=black,colorlinks=true]{hyperref}

\theoremstyle{remark}

\newcommand{\transpose}{\mbox{${}^{\text{T}}$}}

\newcommand{\uvec}[1]{\boldsymbol{\hat{\textbf{#1}}}}

\SetKw{Continue}{continue}
\makeatletter
\newcommand{\removelatexerror}{\let\@latex@error\@gobble}
\makeatother

\DeclareGraphicsExtensions{.pdf,.png,.jpg,.eps}
\IEEEoverridecommandlockouts
\title{\LARGE \bf  Online Vehicle Trajectory Prediction using Policy Anticipation Network\\ and Optimization-based Context Reasoning}
\author{Wenchao Ding and Shaojie Shen%
  \thanks{This work was supported by the Hong Kong PhD Fellowship Scheme (HKPFS). All authors are with the Department of Electronic and Computer Engineering, Hong Kong University of Science and Technology, Hong Kong, China. {\tt\small wdingae@ust.hk, eeshaojie@ust.hk}}%
}
\begin{document}

\maketitle
\thispagestyle{empty}
\pagestyle{empty}

\begin{abstract}
In this paper, we present an online two-level vehicle trajectory prediction framework for urban autonomous driving where there are complex contextual factors, such as lane geometries, road constructions, traffic regulations and moving agents. Our method combines high-level policy anticipation with low-level context reasoning. We leverage a long short-term memory (LSTM) network to anticipate the vehicle's driving policy (e.g., forward, yield, turn left, turn right, etc.) using its sequential history observations. The policy is then used to guide a low-level optimization-based context reasoning process. We show that it is essential to incorporate the prior policy anticipation due to the multimodal nature of the future trajectory. Moreover, contrary to existing regression-based trajectory prediction methods, our optimization-based reasoning process can cope with complex contextual factors. The final output of the two-level reasoning process is a continuous trajectory that automatically adapts to different traffic configurations and accurately predicts future vehicle motions. The performance of the proposed framework is analyzed and validated in an emerging autonomous driving simulation platform (CARLA).
\end{abstract}


\section{Introduction}\label{sec:introduction}
In recent years, there has been growing interest in building fully autonomous vehicles. Our requirement of such vehicles is to have accurate anticipation over other traffic participants so that their planned motions are neither too aggressive nor too conservative. To achieve this goal, autonomous vehicles are expected to reason about the behavior and intentions of surrounding vehicles and subsequently predicts future trajectories of these vehicles.

\begin{figure}[t]
	\centering
	\includegraphics[width = 0.30 \textwidth]{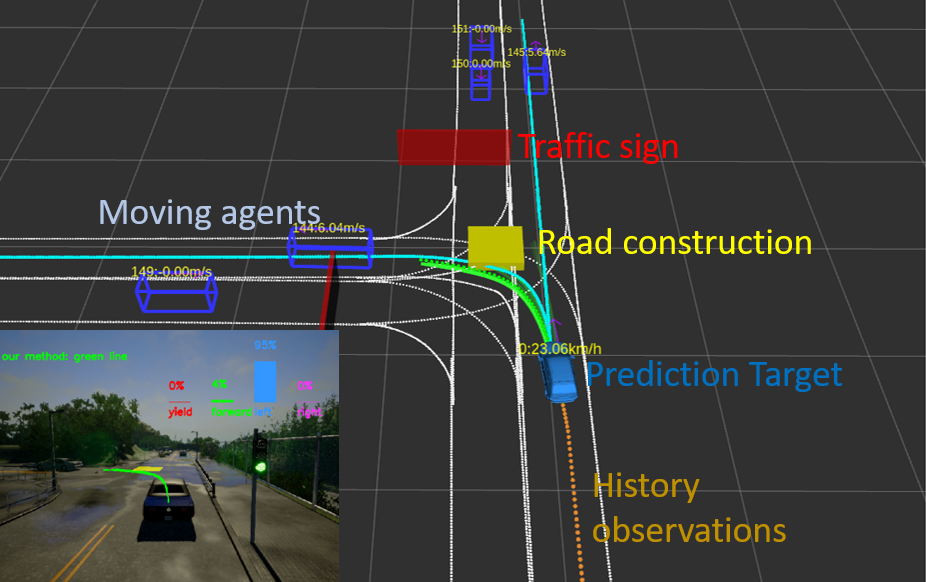}
	\caption{Illustration of the two-level reasoning methodology at an intersection. The two reference lines corresponding to the two possible policies (turn left or go forward) are shown in \textit{cyan}, and the predicted trajectory is shown in \textit{green}. In this example, the high-level policy is first anticipated (namely, turn left) and the relevant contextual information (lane geometry, construction, other agents) is then used in the optimization-based trajectory prediction. More examples can be found in the video \url{https://www.youtube.com/watch?v=r-gSUyFoK8Q}.} \label{fig:motivation_example}
	\vspace{-0.5cm}
\end{figure}

Given an urban driving environment where there are complex latent factors such as lane geometries, traffic regulations, road constructions and dynamical agents, the complexity of the prediction problem is high. Under such a scenario, there are two challenges to be addressed. First, given the complex environment, it is essential to consider the multimodal nature of the future trajectory~\cite{lee2017desire}. For example, at the intersection as depicted in Fig.~\ref{fig:motivation_example}, there are two distinct choices, moving forward and turning left, which result in totally different future trajectories. Second, the prediction method must be highly flexible and able to easily adapt to the complex contextual factors.

Many handcrafted prediction models, such as~\cite{agamennoni2012estimation,laugier2011probabilistic,lefevre2012evaluating,havlak2014discrete}, may lack flexibility and require refactoring when a new contextual factor is introduced. Meanwhile, other methods, especially the popular RNN-based models~\cite{kim2017probabilistic,alahi2016social}, treat the trajectory prediction as a pure regression problem in spite of the multimodal nature of the future trajectory. We are therefore motivated to develop a flexible trajectory prediction framework which can easily adapt to various complex urban environments while incorporating high-level intentions to enhance the prediction accuracy.

In this paper, we propose an \textit{online} two-level vehicle trajectory prediction framework. We develop a policy anticipation network using a long short-term memory (LSTM) network to anticipate high-level policies of vehicles (such as moving forward, yielding, turning and lane changing) based on sequential past observations. Given the high-level policy, we propose an optimization-based context reasoning process in which the complex contextual information is naturally encoded in a multi-layer cost map structure. A policy interpreter is set up to bridge the high-level and low-level reasoning by transforming the policy to a trajectory initial guess of the non-linear optimization. The policy anticipation network is used to capture the intention and guide the trajectory prediction process. Our optimization-based context reasoning process can easily adapt to different traffic configurations by transforming different factors into a unified notation of cost.

The motivation for modeling trajectory prediction as an optimization problem is that human drivers internally balance their maneuvers in terms of the ``cost''. For example, driving through red lights or breaking speed limits would risk receiving penalties, and human drivers have an inborn ability to balance various kinds of costs during driving. The optimization-based reasoning process can be easily extended by adding another cost term to the unified cost map structure.

The idea of modeling drivers as optimizing agents is not new~\cite{wolf2008artificial, abbeel2004apprenticeship, bahram2016combined, sadigh2016planning}, especially in the field of imitating human driving behaviors using inverse reinforcement learning (IRL). However, from the prediction perspective, the multimodal nature of the future trajectory~\cite{lee2017desire,deo2018convolutional} is not well modeled by the optimization process. For example, the non-linear optimization process may converge to either of the two possible intentions in Fig.~\ref{fig:motivation_example}. To this end, we propose the policy anticipation network, which guides the optimization process to the anticipated high-level intention. Note that our optimization-based context reasoning can also incorporate the IRL technique for weight tuning, which is left as important future work.

We summarize the contributions of this paper as follows:
\begin{itemize}
	\item An online two-level trajectory prediction framework which incorporates the multimodal nature of future trajectories.
	\item A highly flexible optimization-based context reasoning process which incorporates a multi-layer cost map structure to encode various contextual factors.
	\item Integration of the vehicle trajectory prediction framework and presentation of the results on accuracy, efficiency, and flexibility in various traffic configurations.
\end{itemize}

The related literature is reviewed in Sect.~\ref{sec:related_works}. A system overview is given in Sect.~\ref{sec:overview}. The main methodology is presented in Sect.~\ref{sec:policy} and Sect.~\ref{sec:optimization}. The implementation details and experimental results are provided in Sect.~\ref{sec:implementation} and Sect.~\ref{sec:results}. Conclusions and future work are given in Sect.~\ref{sec:conclusion}.

\section{Related Works}\label{sec:related_works}
The problem of vehicle trajectory prediction has been actively studied in the literature. As concluded in~\cite{lefevre2014survey}, there are three levels of prediction models, namely, physics-based, maneuver-based and interaction-aware motion models. Physics-based motion models use dynamic and kinematic vehicle models to propagate future states~\cite{ammoun2009real,brannstrom2010model}. However, the prediction results only hold for the very short-term (less than one second). Maneuver-based motion models are more advanced in the sense that the model may forecast relatively complex maneuvers, such as lane change and turns at intersections, by revealing the maneuver pattern. Many of the works on this level present a probabilistic framework to account for the uncertainty and variation of the motion patterns, such as Gaussian processes (GPs)~\cite{tran2014online, laugier2011probabilistic}, Monte Carlo sampling~\cite{eidehall2008statistical}, Gaussian mixture models (GMMs)~\cite{havlak2014discrete} and hidden Markov models~\cite{aoude2012driver}. However, they typically assume vehicles are independent entities and fail to model interactions within the context and with other agents.

Interaction-aware models, on the other hand, take the driving context and vehicle interactions into account, and most of them, such as~\cite{gindele2015learning,lefevre2012evaluating} and~\cite{agamennoni2012estimation}, are based on dynamic Bayesian networks (DBNs). Though these methods are context-aware, they require refactoring the models when considering a new contextual factor. Our method belongs to the interaction-aware level. Compared to the DBN-based prediction methods, our method is more flexible and can be easily adapted to different traffic configurations.

It is notable that recurrent neural networks (RNNs) and their variants, such as LSTM networks, have recently been applied to predict or track moving targets, as in~\cite{kim2017probabilistic,khosroshahi2016surround} and~\cite{ondruska2016deep}. Our policy anticipation network shares a similar structure with~\cite{khosroshahi2016surround}. But the fundamental difference is that the network in~\cite{khosroshahi2016surround} is only used to analyze the maneuver pattern at an intersection and cannot actively predict the future trajectories. Many learning-based end-to-end trajectory prediction models~\cite{kim2017probabilistic, alahi2016social, deo2018convolutional} lack the ability to encode the contextual information. In~\cite{lee2017desire}, Lee~\textit{et al.} suggest combining IRL with an environment feature map to learn the interaction with contextual factors. However, this requires a large amount of training data to generalize due to the high complexity of the model. Also, it is hard to learn the interaction in some rare driving situations, such as red light offences.

\section{System Overview}
\label{sec:overview}
\begin{figure}[t]
	\centering
	\includegraphics[width = 0.35 \textwidth]{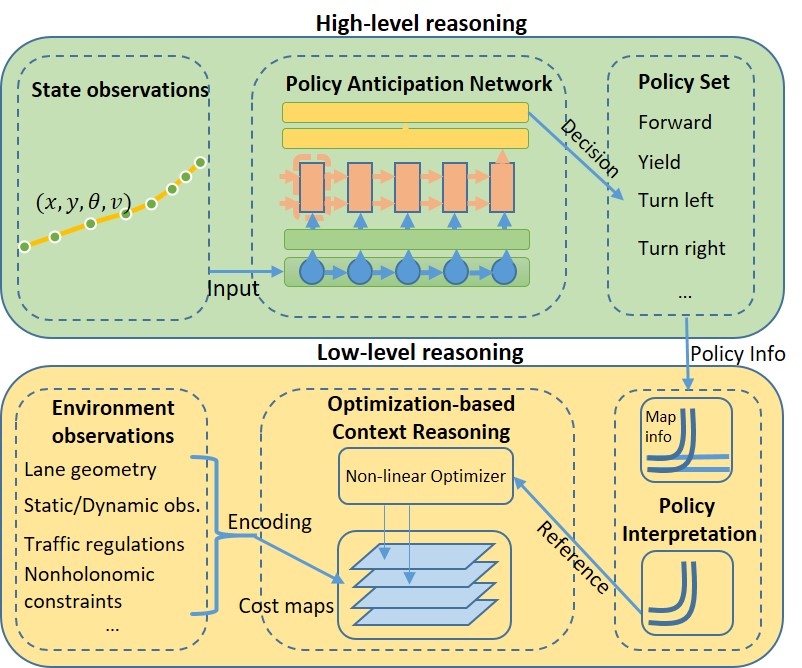}
	\caption{Illustration of the two-level reasoning framework.} \label{fig:framework}
	\vspace{-0.7cm}
\end{figure}

The overview of our vehicle trajectory prediction framework is shown in Fig.~\ref{fig:framework}. During the high-level reasoning, the sequential state observations are fed to the policy anticipation network, which provides the future policy that a vehicle is likely to execute. Together with the map information, the policy can be properly interpreted in the driving context and a reference prediction is generated and fed to the optimization-based context reasoning process. The optimization process renders various environment observations and encodes them into the multi-layer cost map structure. A non-linear optimization process is then conducted to generate the predicted vehicle trajectory.

\section{Policy Anticipation and Interpretation}
\label{sec:policy}
\subsection{Problem formulation}
We assume that the vehicle is equipped with a detector that provides the pose estimation $\mathbf{p}^k_i = (x_i,y_i,\theta_i, v_i)$ of a neighboring vehicle with ID $k$ at different time-instants, where $x_i$ and $y_i$ denote global coordinates at frame $i$,  $\theta_i$ denotes the vehicle orientation in the 2-D plane, and $v_i$ denotes the body velocity. We accumulate observations from different time-instants inside a sliding window with a total window size of $T_{\text{obs}}$. And the network predicts vehicle $k$'s future policy in a look-ahead window from $T_{\text{obs}}$ to $T_{\text{pred}}$. The annotated labels include \textit{forward}, \textit{yield}, \textit{turn left}, \textit{turn right}, \textit{lane change left}, and \textit{lane change right}, and the labels can be easily extended when considering complex lane geometries.

\subsection{Network structure}
Our policy anticipation network is based on an RNN encoder structure~\cite{cho2014rnnecdc}. We refer interested readers to~\cite{cho2014rnnecdc} and~\cite{lee2017desire} for the detailed structure. Note that the output layer is modified to a softmax layer to provide the likelihood for all the policy labels. The probability distribution is used in the interpretation of the policy in Sec.~\ref{sec:policy_interpretation}. We adopt negative log-likelihood (NLL) loss for this classification problem.

\subsection{Policy interpretation}
\label{sec:policy_interpretation}
The policy interpretation module combines the policy anticipation results with a local map, so that the optimization-based context reasoning can start with a reasonable initial guess.
\begin{figure}[h]
	\centering
	\includegraphics[width = 0.41 \textwidth]{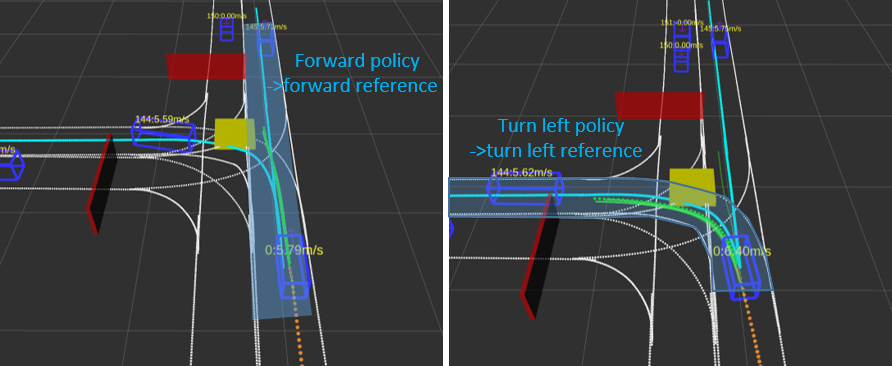}
	\caption{Illustration of the policy interpretation. The local reference lines extracted from the map are marked in \textit{cyan}. The figure on the left illustrates that when the vehicle (ID 0) has not shown any intention to turn, it is anticipated to be executing a ``forward'' policy, so the forward reference line is extracted. After the vehicle shows a left-turn pattern, the left-turn reference line is extracted. The local context region is marked in the transparent cyan area.}
	\label{fig:policy_interp}
\end{figure}
As shown in Fig.~\ref{fig:policy_interp}, with different initial guesses (turning left or forward in this case), the optimization will be devoted to finding a solution in a totally different local solution space. Specifically, we use the likelihood provided by the policy anticipation network as follows: 1) we prune the infeasible anticipations (turning right in this example); 2) we take the policy of the maximum likelihood, and 3) we generate an initial trajectory prediction by extracting reference points corresponding to the selected policy. The initial guess is fed to the optimization-based context reasoning for further processing. In the future, instead of using deterministic reasoning based on one selected policy, we plan to use a probabilistic interpretation process.

\section{Optimization-based Context Reasoning}
\label{sec:optimization}
\subsection{Cost map structure} \label{sec:cost_map_design}
\begin{figure}[h]
	\centering
	\includegraphics[width = 0.35 \textwidth]{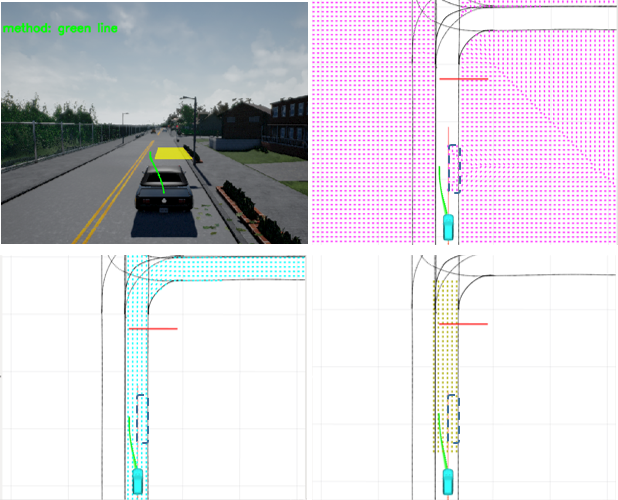}
	\caption{Illustration of the multi-layer cost map structure. The top-left image is captured from CARLA, and a road construction site is marked in \textit{yellow}. The top-right figure shows the static layer with repulsive forces (cost) pointing to the free space. The bottom-left image illustrates the costs induced by the desired velocity, and the bottom-right image shows the cost induced by the red light. Different forces may be conflicting (as around the dashed box).}\label{fig:multi_layer}
	\vspace{-0.4cm}
\end{figure}

In this section, we present the cost map structure, which encodes the whole driving context. We specify different kinds of costs by separating them into different layers with distinct physical meanings, for the sake of illustration. A toy example of the multi-layer cost map is given in Fig.~\ref{fig:multi_layer}. We adopt a four-layer cost map design in which we encode the cost induced by the lane geometry and static obstacles into the \textit{static layer}, the cost induced by the moving objectives (MO) into the \textit{MO layer}, the cost induced by traffic regulations into the \textit{context layer}, and the cost induced by the vehicles' nonholonomic constraints into the \textit{nonholonomic layer}.

\subsection{Cost functions}
We adopt a discrete notation of the vehicle trajectory~\cite{ziegler2014trajectory} where a continuous trajectory $\mathbf{x}(t) =  \left( x(t), y(t) \right)\transpose $ is represented by a series of rear axle center points $\mathbf{x}_i =  \left( x_i, y_i \right)\transpose $  in a global coordinate system. Namely, the predicted trajectory is approximated by $N$ points $\mathbf{x}_i = \mathbf{x}(t_i)$, which are sampled at equidistant times $t_i = t_0 + ih, 0\leq i < N$ of sampling step width $h$. The dynamics of the trajectory $\mathbf{x}(t)$ can be expressed as a function of its time derivatives, which are the finite differences of the sampling points. The orientation and curvature of the trajectory can be expressed by its time derivatives~\cite{ziegler2014trajectory}.
Following these notations, we introduce the cost functions $f(\mathbf{x})$.

\begin{figure}[t]
	\centering
	\includegraphics[width = 0.30 \textwidth]{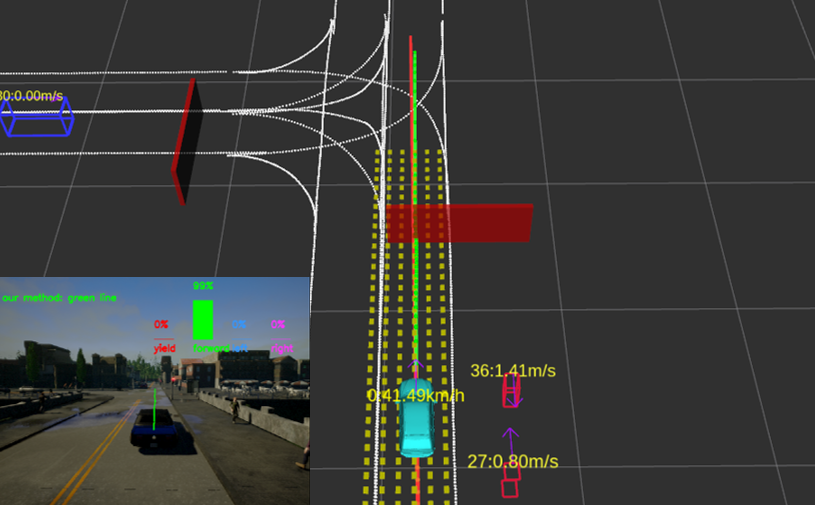}
	\caption{Illustration of red light offence. When the kinetic energy of the vehicle can overcome the artificial repulsive force induced by the red light, the red light offence is captured and a warning is provided by our prediction system.}\label{fig:through_red}
	\vspace{-0.9cm}
\end{figure}

\begin{enumerate}[wide, labelwidth=!, labelindent=9pt]
	\item \textit{Lane geometry}: Ideally, the point $\mathbf{x}$ that exceeds the solid-lane boundary should receive a repulsive force (cost) $f_g(\mathbf{x}) $ pointing into the travelable lanes. For broken-lane boundaries, we pose the cost of the same structure, but the magnitude is much smaller to allow for lane changing. We present a bi-directional signed distance field (bi-SDF) to describe the corresponding cost characteristics:
	\begin{equation}
		f_g( \mathbf{x} ) =
			\begin{cases}
					\alpha_g (- d_b(\mathbf{x}) + \tau_b) ^2 &\text{if $ d_b(\mathbf{x}) \leq 0$}
					\\
					\alpha_g (  d_b(\mathbf{x}) - \tau_b) ^2 &\text{if $0 < d_b(\mathbf{x})\leq \tau_b$}\\
					0 &\text{if $\tau < d_b (\mathbf{x})$},
			\end{cases}
	\end{equation}
	where $d_b (\mathbf{x})$ measures the distance to the nearest solid-lane boundary, $\tau_b$ is the distance threshold, and $\alpha_g$ is the cost magnitude. Note that $d_b(\mathbf{x}) > 0$ means the in-boundary area, while $d_b(\mathbf{x})<0$ represents that the point exceeds the boundary and needs to be pushed back to the travelable lanes. Different from the traditional SDF, which does not define the gradient when $d_b (\mathbf{x}) < 0$, we slightly extend the definition so that the point outside of the boundary will receive a force pointing inside the lane. The benefit of extending $d_b (\mathbf{x}) < 0$ is that the optimization process is less likely to get stuck in the infeasible out-of-boundary area.
	\item \textit{Static obstacles/ driveable area}: The cost $f_s(\mathbf{x})$ induced by static obstacles shares a similar form to $f_g(\mathbf{x})$, and these two costs are categorized into the static layer. The distance measure to static obstacles $d_s(\mathbf{x})$ is also extended to allow a negative distance.
	\item \textit{Moving obstacles}: To take interaction with other agents into account, we introduce a cost $f^j_d(\mathbf{x}_i)$ for $t_i$ if the position $\mathbf{x}_i$ of the predicted vehicle is within a distance threshold $\tau_d$ of the prediction $\mathbf{x}_{\text{pred},i}^{j} $ of another agent $j \in \mathcal{J}$, where $\mathcal{J}$ denotes the set of all the interacting agents. The practical method of acquiring $\mathbf{x}_{\text{pred},i}^{j} $ is introduced in Sec.~\ref{sec:optimization_procedure}. The MO cost at time $t_i$ is given by
	\begin{equation}
		f^j_d(\mathbf{x}_i) = \alpha_d (d_o(\mathbf{x}_i, j) - \tau_o)^2	\mathbb{1}_{d_o(\mathbf{x}_i,j) < \tau_o},
	\end{equation}
	where $f^j_d(\mathbf{x}_i)$ is specified by the quadratic error between the distance $d_o(\mathbf{x}_i,j)$ to the moving agent $j$ and $\tau_0$ if the distance threshold $\tau_0$ is reached.
	\item \textit{Red lights}: We argue that red lights should not be enforced as hard constraints since in a real-world driving scenario there exist red light offences. To capture the real intention of other drivers under traffic control, we introduce a red light repulsive force $r(\mathbf{x})$. The repulsive force is supposed to produce larger resistance for vehicles travelling at higher velocity. It is notable that if a vehicle refuses to brake and tries to go through a red light, as shown in Fig.~\ref{fig:through_red}, the cost $f_r(\mathbf{x})$ will not dominate the optimization process and the abnormal behavior is captured. The overall cost $f_r(\mathbf{x})$ can be expressed by the dot product of the velocity $\dot{\mathbf{x}} $ and the repulsive force $r (\mathbf{x})$ as follows:
	\vspace{-0.2cm}
	\begin{equation}
		\begin{aligned}
		f_r(\mathbf{x}) 	& = \norm{ \dot{\mathbf{x}} \cdot r (\mathbf{x}) }^2 \\
							& =
							\begin{cases}
								\alpha_r (d_r(\mathbf{x})-\tau_r ) ^2  \norm{\dot{\mathbf{x}} \cdot \uvec{r}(\mathbf{x})}^2 &\text{if $0 < d_r(\mathbf{x})\leq \tau_r$}\\
								0 &\text{if $\tau_r < d_r (\mathbf{x})$},
							\end{cases}
		\end{aligned}
	\end{equation}
	where $\alpha_r$ is the cost magnitude, $\uvec{r}(\mathbf{x})$ denotes the unit direction of the force $r(\mathbf{x})$, $d_r (\mathbf{x})$ denotes the distance to the red light, and $\tau_r$ is the distance threshold below which the force $r(\mathbf{x})$ will take effect.

	\item \textit{Speed limits}: Like red lights, speed limits should not be encoded in hard constraints when taking speed limit offences into account. We introduce the cost $f_v(\mathbf{x})$, which is induced by the speed limit and should also allow the vehicle to stop in the case of a traffic jam. As a result, we model $f_v(\mathbf{x})$ as the quadratic error between the predicted velocity $\dot{\mathbf{x}}$ and a desired velocity $\mathbf{v}_{\text{des}}$. The magnitude of the desired velocity $\norm{ \mathbf{v}_{\text{des}}} $ is determined by the minimum between two factors, namely, the speed limit $v_{\max}$ and the velocity trend $v_{\text{trend}}$. Specifically, $v_{\text{trend}}$ is obtained by conducting velocity fitting for the historical velocity observations in $T_{\text{obs}}$, which captures the acceleration and deceleration trend of the predicted vehicle and is close to zero in the case of a traffic jam. The direction of the desired velocity $\uvec{v}_{\text{des}}$ conforms to the lane geometry. Mathematically, we have $\norm{\mathbf{v}_{\text{des}}} = \min {(v_{\max}, v_{\text{trend}}) } $ and $f_v(\mathbf{x}) = \norm{\dot{\mathbf{x}} - \mathbf{v}_{\text{des}} }^2$.

	\item \textit{Nonholonomic constraints}: The predicted trajectory should obey the limits of the vehicle motion model. Due to the steering geometry of the vehicle, the curvature should be bounded by the maximum curvature allowed. However, when taking abnormal operations, such as skidding, into account, the hard curvature constraint should also be modeled by the feasibility cost $f_{\kappa}(\mathbf{x})$ as follows:
	\vspace{-0.1cm}
	\begin{equation}
		f_{\kappa}(\mathbf{x}) =\alpha_{\kappa}  (\kappa(\mathbf{x}) - \kappa_{\max})^2 \mathbb{1}_{\kappa(\mathbf{x}) > \kappa_{\max}},
	\end{equation}
	where the cost takes effect when the curvature exceeds the limit $\kappa_{\max}$ and $\alpha_{\kappa}$ is the cost magnitude. Similarly, due to the friction limit of tires and throttle limit of vehicles, the maximum acceleration of vehicles cannot exceed a limit $a_{\max}$. We model the acceleration feasibility cost $f_{a}(\mathbf{x})$ as follows:
	\vspace{-0.1cm}
	\begin{equation}
		f_{a}(\mathbf{x}) =\alpha_{a} (\ddot{\mathbf{x}} - a_{\max})^2\mathbb{1}_{\ddot{\mathbf{x}} > a_{\max}},
	\end{equation}
	where the maximum acceleration is denoted by $a_{\max}$ and cost magnitude is denoted by $\alpha_{a}$.
\end{enumerate}

The motivation for using quadratic functions with barriers for the cost functions is that 1) they tolerate a mild deviation from the best driving practices, and 2) they penalize abnormal behaviors while still allowing their existence.

\subsection{Non-linear optmization}\label{sec:optimization_detail}
Based on the cost functions, we now introduce the optimization formulation. At a top level, the predicted trajectory is generated by minimizing $J(\mathbf{x}(t))$, which is the integral of the overall loss $ L(\mathbf{x}(t))$ over time $T$, i.e., $J(\mathbf{x}(t)) = \int_{t_0}^{t_0 + T} L(\mathbf{x}(t))$, which can be approximated by finite summation in the discrete case as follows:
\begin{equation}
	\label{eq:non-linear-opt}
	\begin{aligned}
		J(\mathbf{x}(t)) = &\sum_{i=0}^{N-1} \big(  w_g f_g(\mathbf{x}_i) +  w_s f_s(\mathbf{x}_i)+w_d\sum_{j\in \mathcal{J}} f^j_d(\mathbf{x}_i) \\
		+& w_r f_r(\mathbf{x}_i) + w_v f_v(\mathbf{x}_i)+w_{\kappa} f_{\kappa}(\mathbf{x}_i) + w_a f_a(\mathbf{x}_i) \big) h.
	\end{aligned}
\end{equation}

The weights of different costs represent the tradeoff among different contextual factors. We tune the weights so that predicted trajectories match a human prior for different traffic configurations. As mentioned in Sec.~\ref{sec:introduction}, the optimization process can incorporate IRL for automatic weight tuning, which is important future work.

\section{Implementation Details} \label{sec:implementation}
\subsection{Simulation environment} \label{sec:simulation_environment}
We adopt an open-source urban autonomous driving simulator named CARLA~\cite{Dosovitskiy17}. In this section, we present our environment setup. For a scene containing $n$ vehicles, the first $n-2$ vehicles (agent vehicles) are controlled by the autopilot module provided by CARLA, the $n-1$-th vehicle (player vehicle) is controlled by a human player and the $n$-th vehicle is an observer vehicle which is supposed to closely follow the player vehicle, sense the environment, and predict the trajectory of the player vehicle. We focus on predicting the trajectories for the player vehicle since it reflects real human intentions. Another reason is that the agent vehicles do not have complex maneuver patterns due to the fixed handwritten logic of the autopilot module. Hence, when presenting the experimental results (Sec.~\ref{sec:results}) we will focus on illustrating the prediction results for the player vehicle, to give a clean and informative visualization.

\subsection{Data collection and network training}
We collect the training data for the policy anticipation network from CARLA by driving the player vehicle ourselves using a Logitech G29 racing wheel. During the driving, we follow the traffic rules most of the time and conduct different maneuver patterns, but we also commit intentional traffic rule offences, as in Fig.~\ref{fig:through_red}, to examine how our prediction module will respond. Moreover, we add virtual road construction sites, as in Fig.~\ref{fig:multi_layer}, and respond to them during driving using the feedback from our visualization system. The collected data is $21,260$ frames in total. $T_{\text{obs}}$ and $T_{\text{pred}}$ are both set to 40 frames (4s). The policy label can be determined by examining the statistics on the steering angle and acceleration in the $T_\text{pred}$ in an unsupervised way. One problem with the data collected from CARLA is that the current version\footnote{CARLA release 0.7.0 is used for all the experiments.} only includes two-lane roads with traffic moving in opposite directions, which means that lane change behavior cannot be effectively incorporated. In the future, we will collect data from more complex environments to enrich the dataset.


\subsection{Non-linear optimization procedure}
\label{sec:optimization_procedure}
The non-linear optimization formulation~(\ref{eq:non-linear-opt}) is implemented in Ceres~\cite{ceres-solver} since the objectives can be rewritten into non-linear least squares. If more complex objectives are involved, non-linear solvers such as NLOPT~\cite{Johnson2011} can be used. The maximum number of iterations is set to $20$. Recall that the prediction for a certain vehicle can depend on the prediction of other vehicles due to the moving obstacle cost term $f^j_d(\mathbf{x})$. In practice, we use the prediction results from the last prediction round to calculate $f^j_d(\mathbf{x})$.

\section{Results}
\label{sec:results}
\subsection{Prediction accuracy}
\label{sec:result_accuray}
\begin{figure*}[t]
	\centering
		\begin{subfigure}[b]{0.245\textwidth}
			\includegraphics[trim=-1.5cm -0.0cm -0.5cm -1.0cm, width =\textwidth]{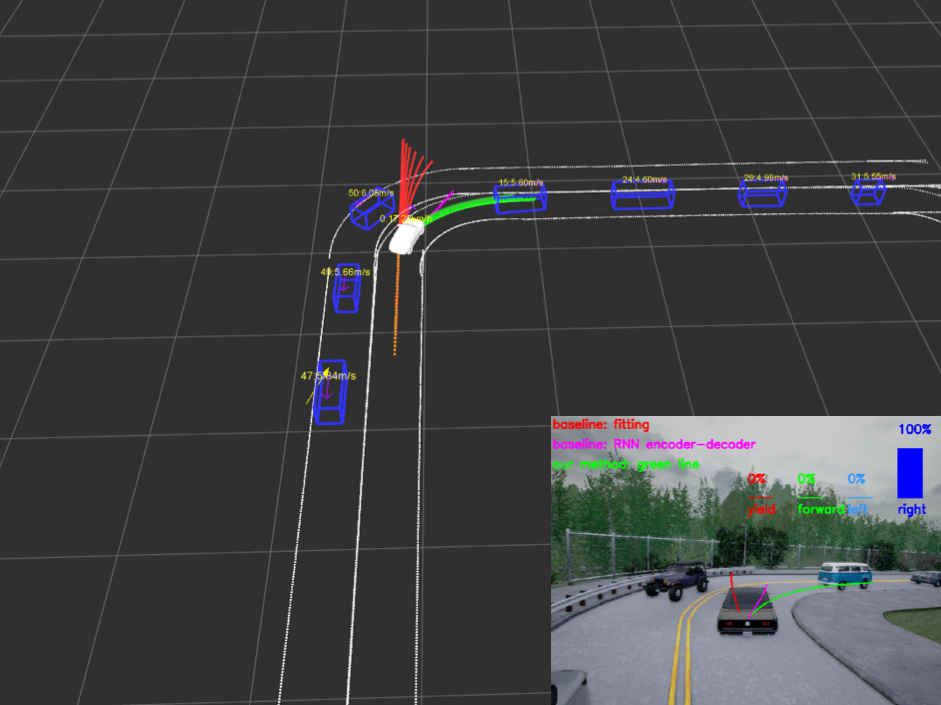}
			\includegraphics[width =\textwidth]{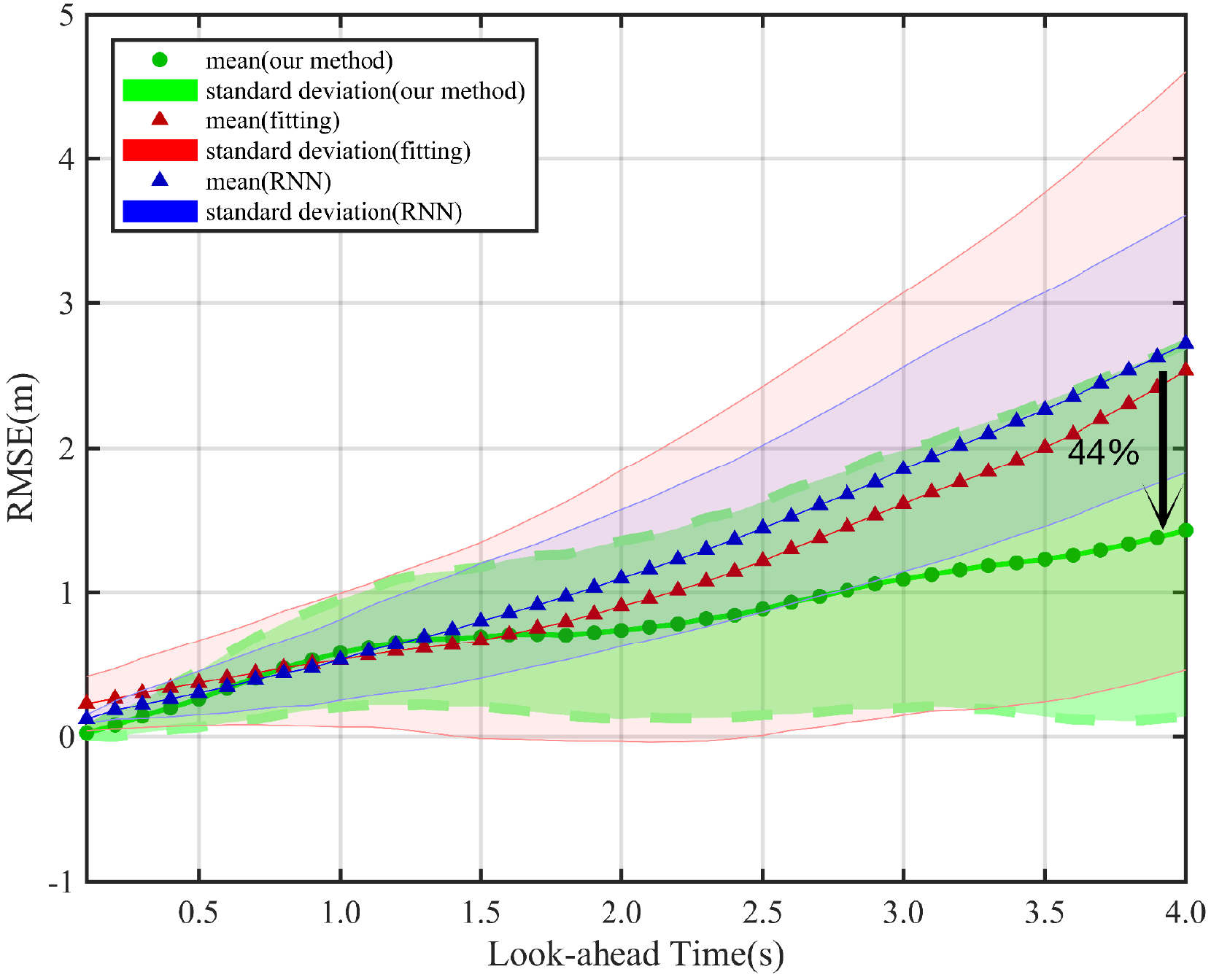}
			\includegraphics[width =\textwidth]{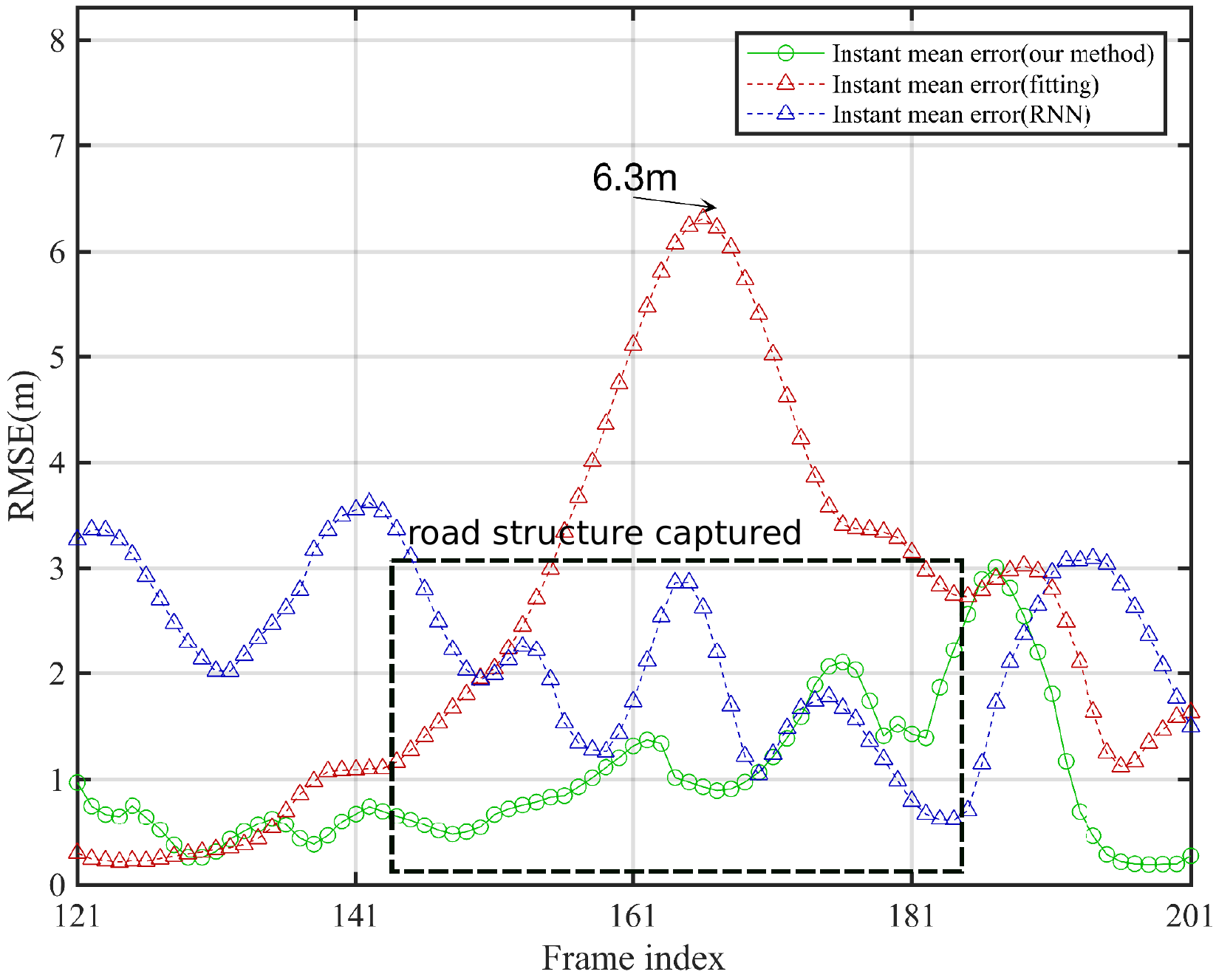}
			\caption{Curved road}\label{fig:simu_curvy_road}
		\end{subfigure}%
		\begin{subfigure}[b]{0.245\textwidth}
			\includegraphics[trim=-1.5cm -0.0cm -0.5cm -1.0cm, width =\textwidth]{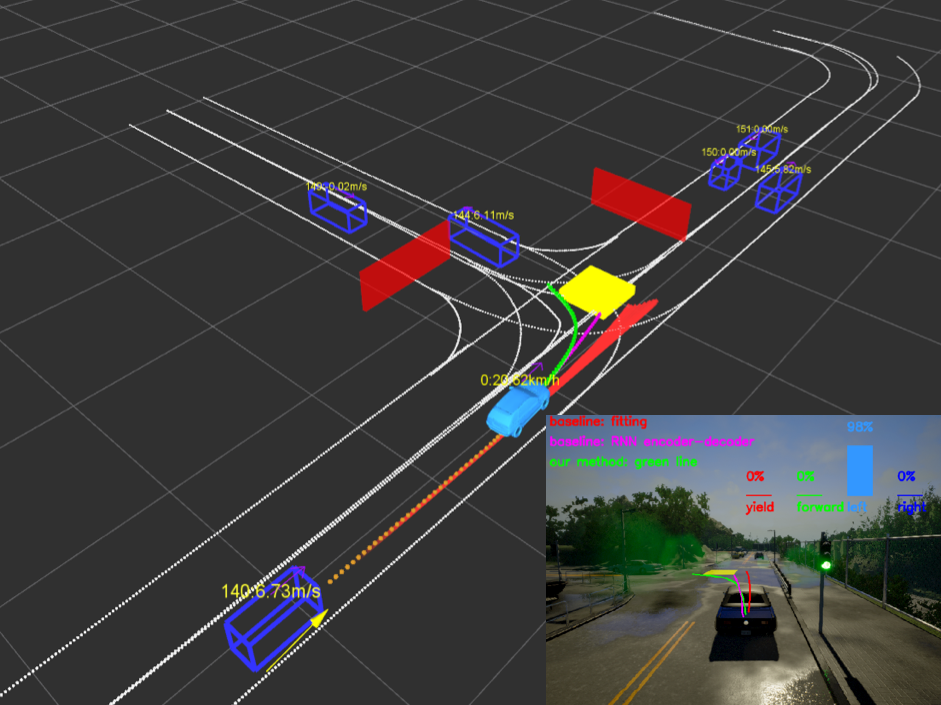}
			\includegraphics[width =\textwidth]{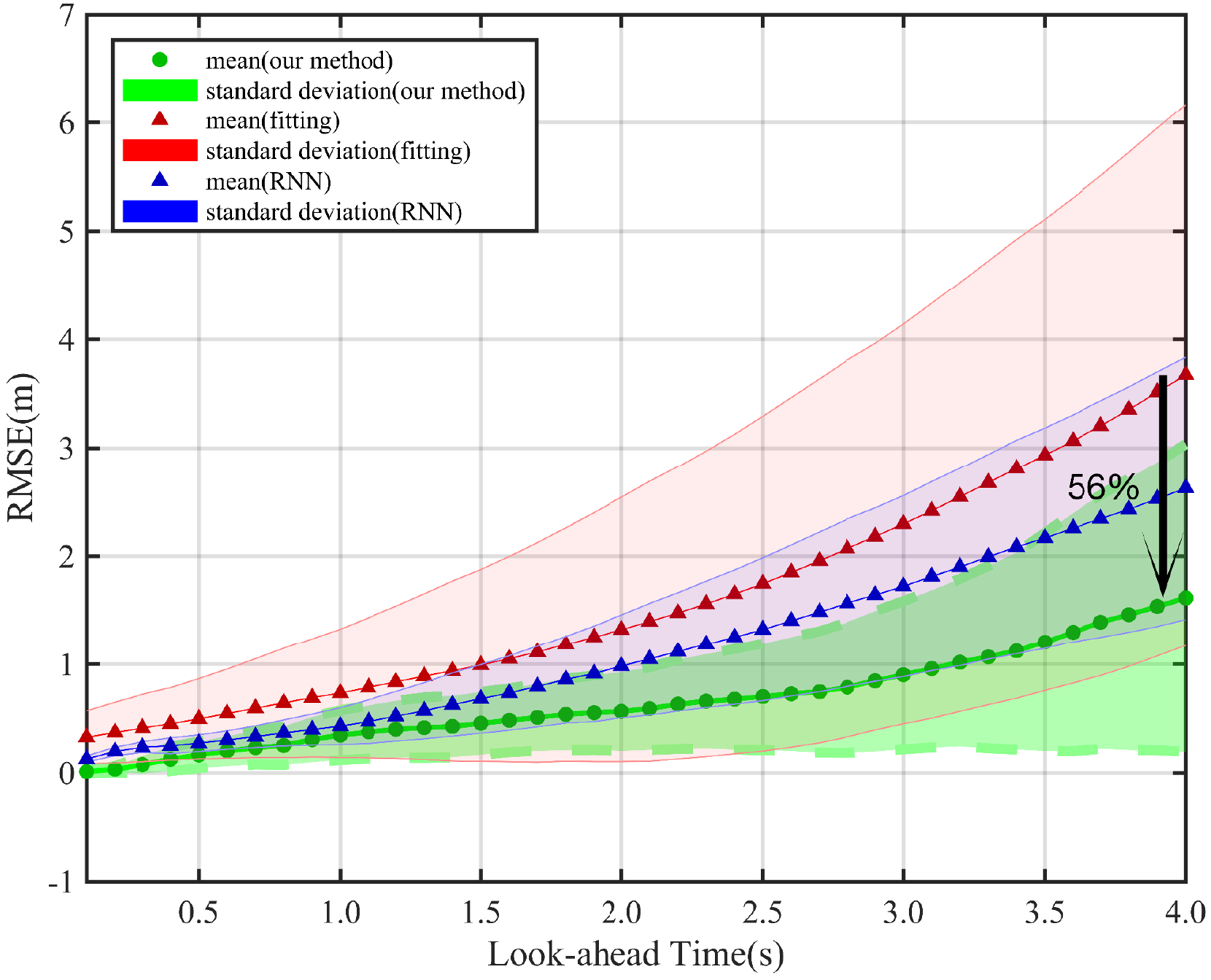}
			\includegraphics[width =\textwidth]{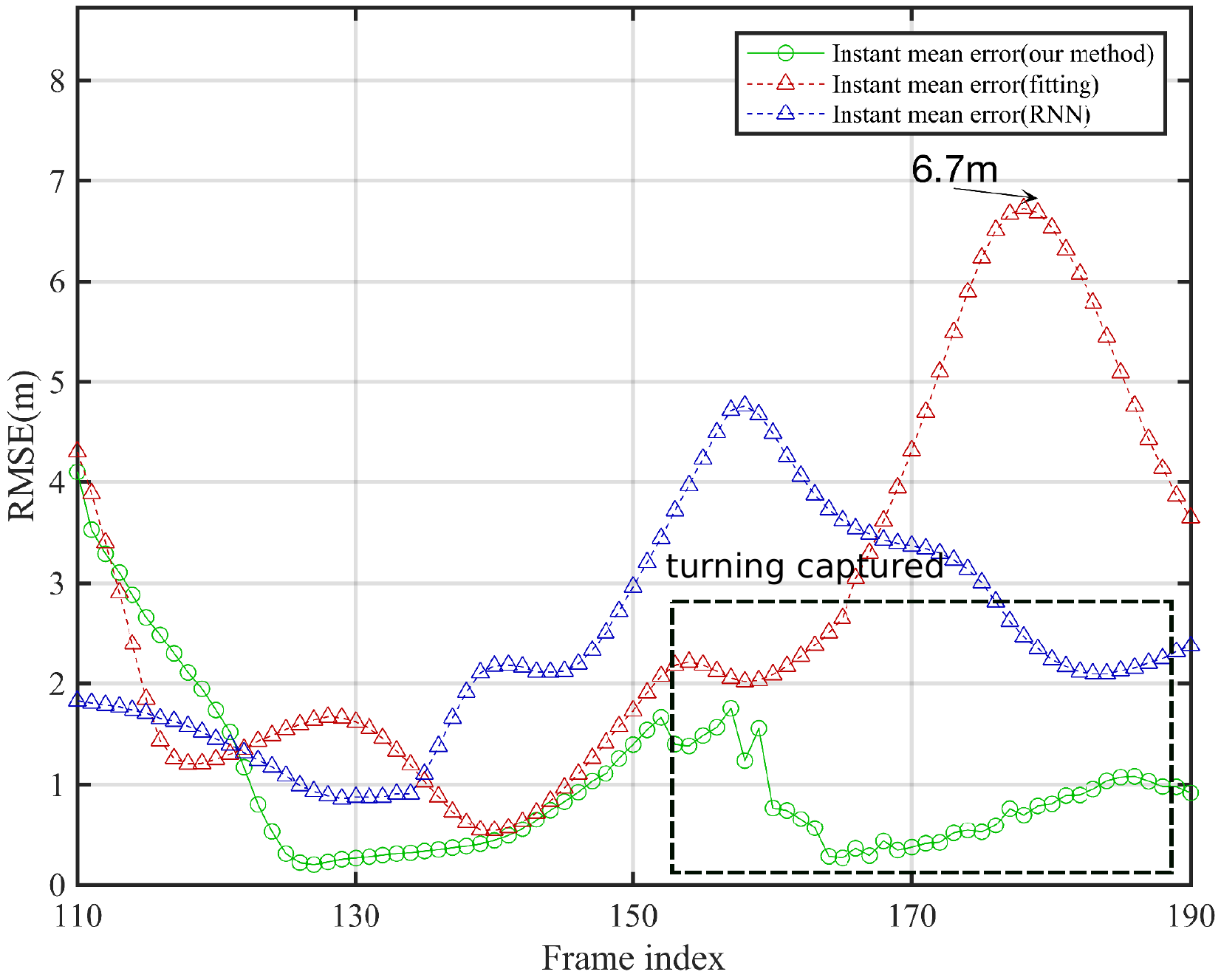}
			\caption{Intersection}\label{fig:simu_int}
		\end{subfigure}%
		\begin{subfigure}[b]{0.245\textwidth}
			\includegraphics[trim=-1.5cm -0.0cm -0.5cm -1.0cm, width =\textwidth]{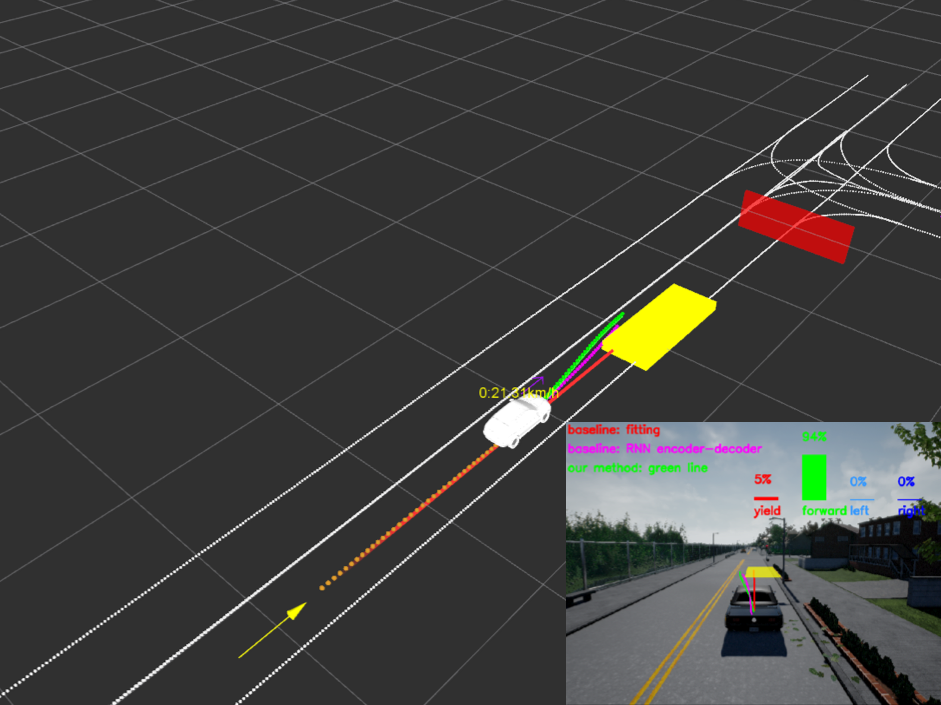}
			\includegraphics[width =\textwidth]{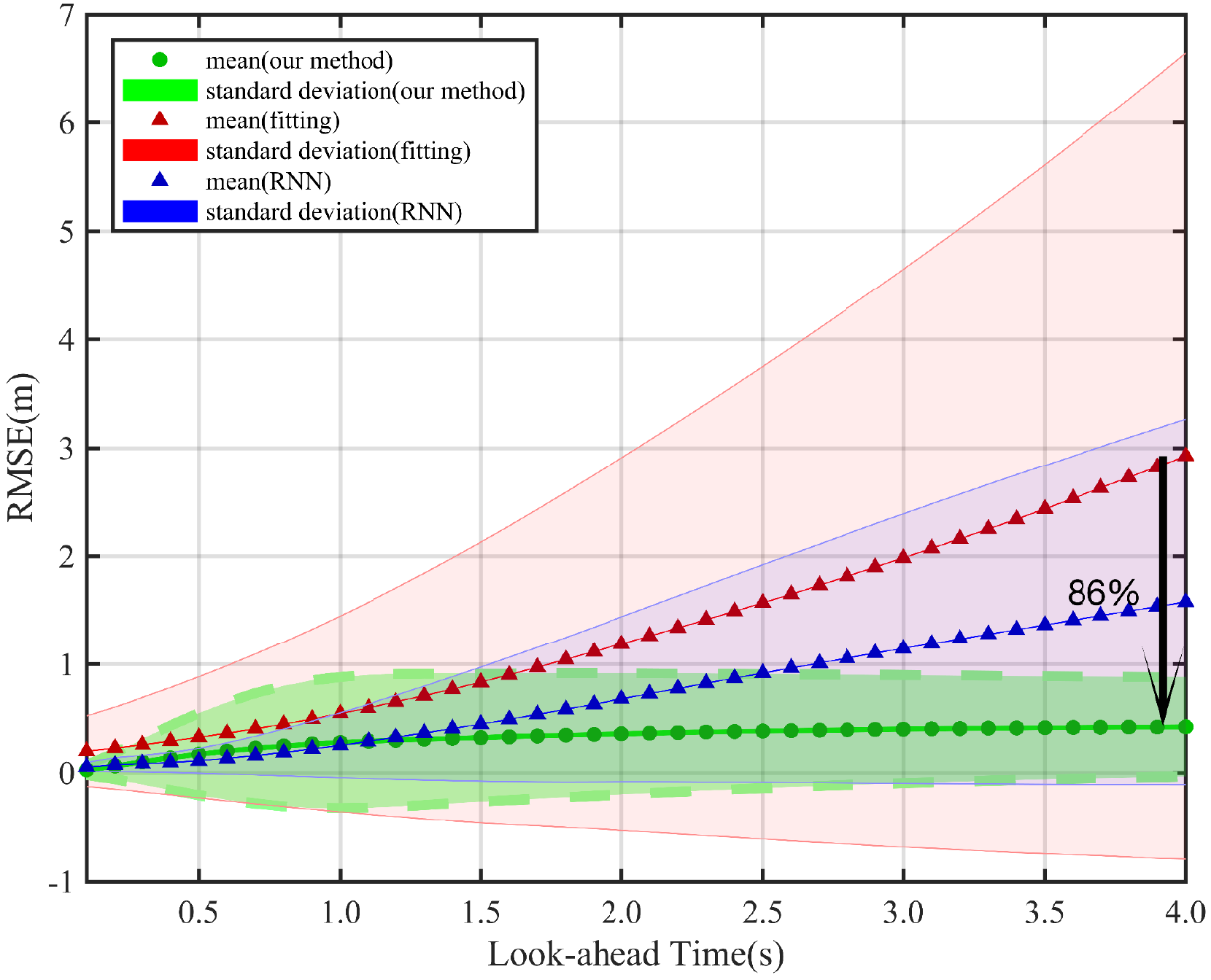}
			\includegraphics[width =\textwidth]{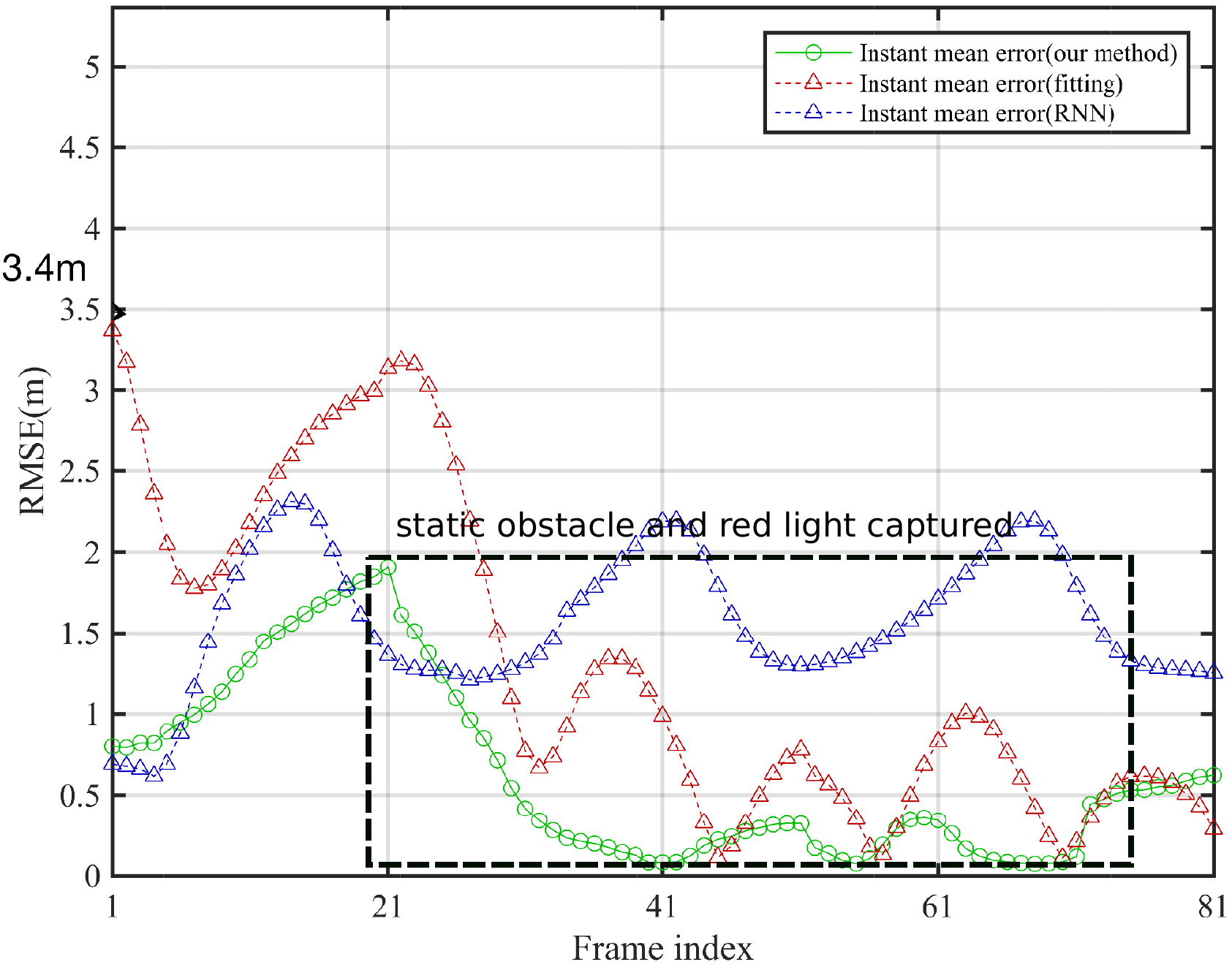}
			\caption{Traffic control}\label{fig:simu_light}
		\end{subfigure}%
		\begin{subfigure}[b]{0.245\textwidth}
			\includegraphics[trim=-1.5cm -0.0cm -0.5cm -1.0cm, width =\textwidth]{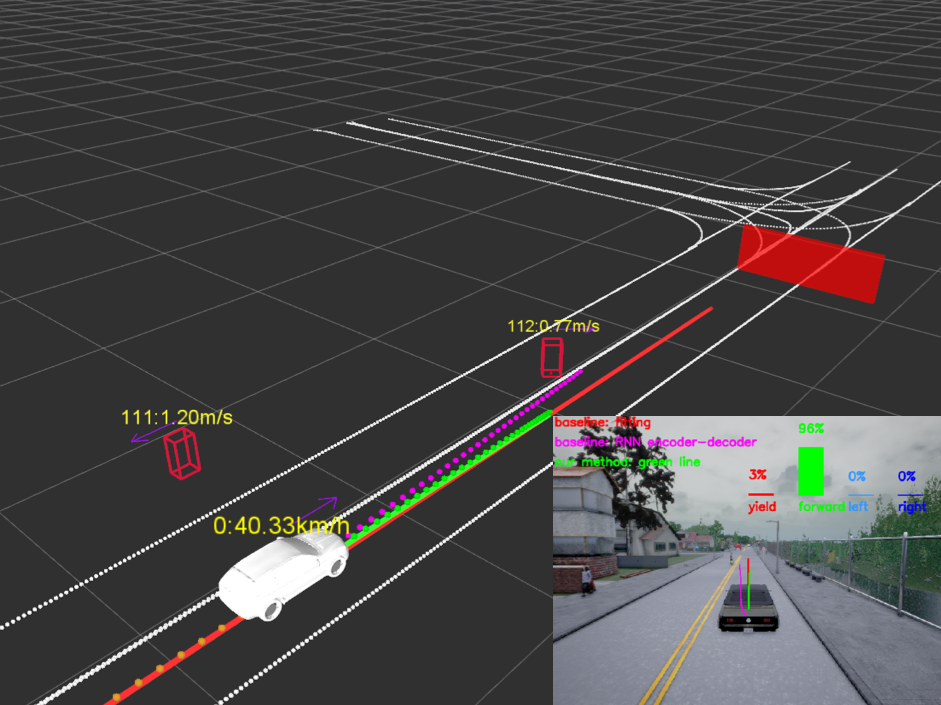}
			\includegraphics[width =\textwidth]{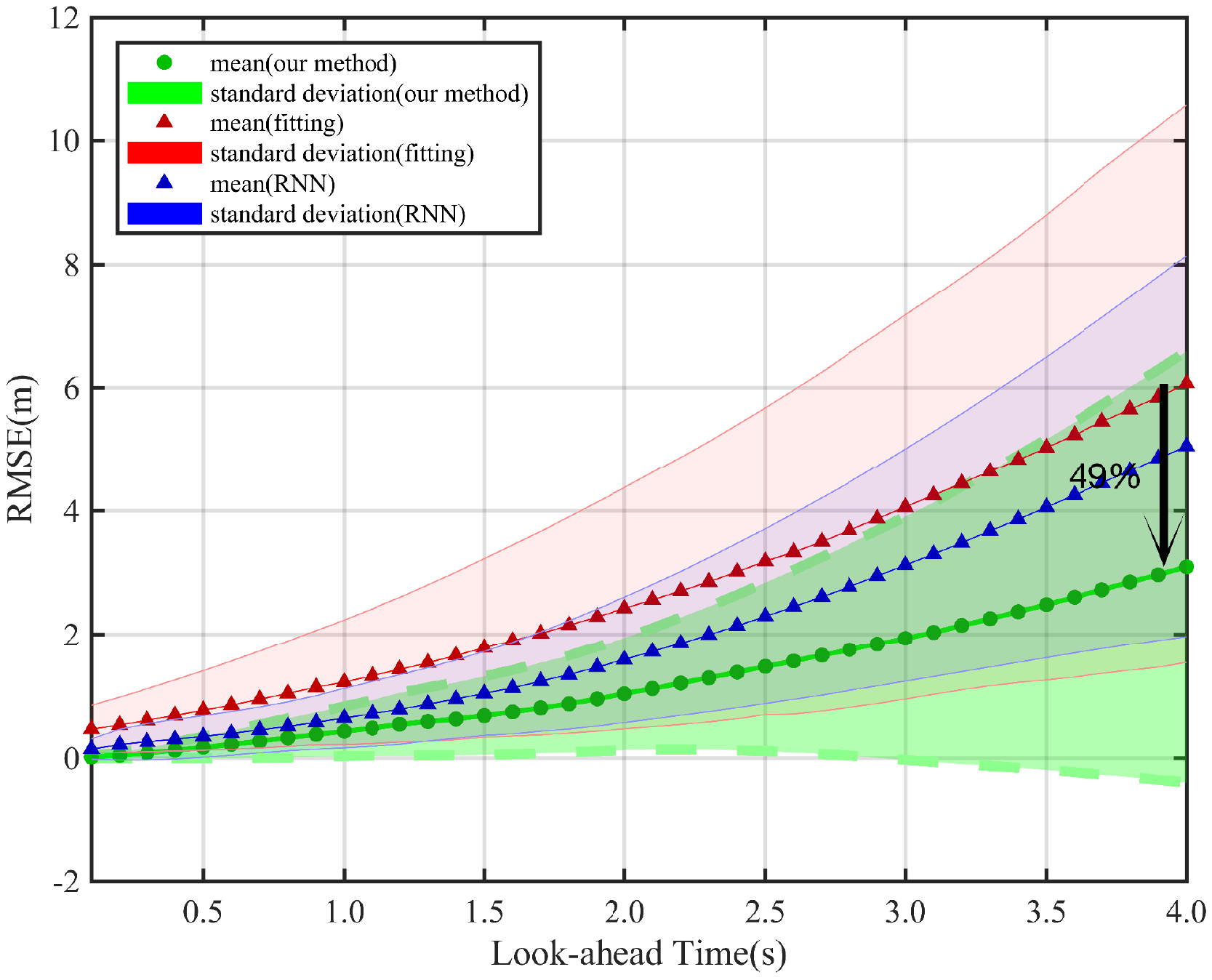}
			\includegraphics[width =\textwidth]{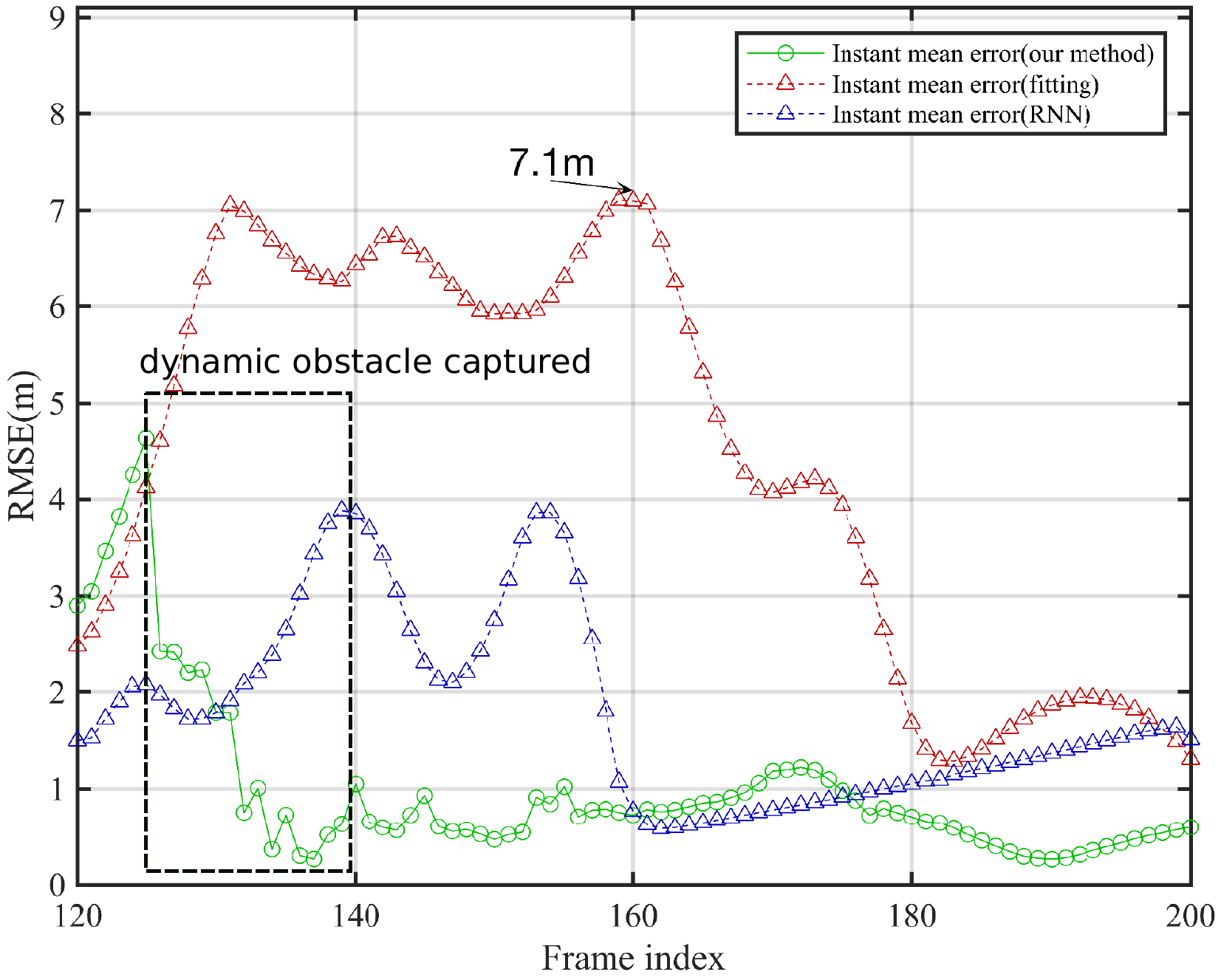}
			\caption{Moving obstacles}\label{fig:simu_dno}
		\end{subfigure}%
	\caption{Illustration of the comparsion in different driving scenarios. The top row of figures visualizes the environment together with the image from the virtual sensor. The vehicles are represented by \textit{blue} bounding boxes with their IDs and velocities on the top. The past observations are marked in \textit{orange}, and the prediction results are marked in \textit{green} (ours),~\textit{red} (fitting) and \textit{purple} (RNN) for our proposed method and the two baselines, respectively. The middle row of figures shows the error statistics during the entire test cases (the semantics are elaborated in Sec.~\ref{sec:result_accuray}). The bottom row of figures illustrates the instantaneous average prediction error over the prediction horizon of each frame in the region of interest (about $4$ s around the event).}
	\vspace{-0.2cm}
	\label{fig:runtime_sim}
\end{figure*}

We adopt the root mean square error (RMSE) between the predicted coordinates and the true coordinates as the error metric. We are concerned with how the RMSE error statistics change with respect to the look-ahead time, especially when the look-ahead time is large. To this end, we plot the mean and variance of the RMSE loss with respect to the look-ahead time, as shown in Fig.~\ref{fig:runtime_sim}. We compare our method with the following two methods:
\begin{itemize}
	\item \textit{Naive fitting method}. The future trajectory is generated using least mean square polynomial regression with an acceleration regulator. This method can capture the trend but cannot incorporate the driving context.
	\item \textit{RNN encoder-decoder trajectory regression}. This method uses an RNN to encode the past maneuver history and directly outputs the future trajectory through the RNN decoder. This structure is popular, and is adopted in~\cite{lee2017desire} and~\cite{deo2018convolutional}.
\end{itemize}
Since the source code of~\cite{lee2017desire} is not officially available and~\cite{deo2018convolutional} is mainly tested in a highway dataset, we adopt the RNN encoder-decoder part in~\cite{lee2017desire} according to the available implementation details~\cite{lee2017sup}. We conduct the experiments in the form of case studies to show that our proposed framework can easily adapt to various traffic configurations, as elaborated in Sec.~\ref{sec:simulated_cases}.

\subsection{Testing in different traffic configurations}
\label{sec:simulated_cases}
To verify that our proposed method can automatically adapt to different traffic configurations and take various latent factors into account, we design five test cases: driving along a curved road, heading towards a pedestrian who is crossing the road, passing through an intersection with road construction, heading towards a red light with road construction, and committing a red light offence. To give a clean visualization, we focus on the prediction for the vehicle being driven by us, namely, the vehicle with ID $0$.

\begin{enumerate}[wide, labelwidth=!, labelindent=9pt]
\item \textit{Curved road}:
This case is used to verify the capability of reasoning about lane geometries. As illustrated in Fig.~\ref{fig:simu_curvy_road}, both baseline methods can capture the motion trend. However, because they are unaware of the lane geometries, they take a long time to conform to the shape of the road. On the other hand, our proposed method produces a reasonable prediction immediately. Quantitatively, our method achieves $44\%$ accuracy improvement for the ending frame in $T_{\text{pred}}$. From the instantaneous error statistics, i.e., the average error for the whole predicted trajectory, we observe that the maximum instantaneous error is reduced from $6.3$ m to $3$ m. This testing case verifies the effectiveness of optimization-based context reasoning.

\item \textit{Intersection with road construction}:
This case is used to illustrate the importance of high-level reasoning, which the two baseline methods lack. As shown in Fig.~\ref{fig:simu_int}, neither baseline methods can effectively capture the turning left intention and both converge slowly. The benefit of incorporating high-level behavior anticipation is validated by an accuracy gain of $56\%$ for the ending frame and a lower instantaneous error during the intersection entrance. The results verify that it is essential to incorporate the high-level intention.

\item \textit{Red light with road construction}:
This case is taken as one example of the non-linear optimization (Fig.~\ref{fig:multi_layer}). The statistics are provided in Fig.\ref{fig:simu_light}, which confirms the necessity of modeling contextual factors.

\item \textit{Heading towards a pedestrian}:
This case is used to illustrate the ability to reason about other moving agents. As shown in Fig.~\ref{fig:simu_dno}, the predicted vehicle is moving at high speed, but a pedestrian is crossing the road ahead of the vehicle. Sudden braking of the vehicle should be the reaction. As shown in Fig.~\ref{fig:simu_dno}, the two baselines are still giving out forward trajectories, while our method expects hard braking by modeling the interactions between agents. The instantaneous error shows that our method predicts the braking intention beforehand.

\item  \textit{Red light offence}:
This case is used to show how the proposed method responds to abnormal driving behavior, and is elaborated in Fig.~\ref{fig:through_red}.
\end{enumerate}

\subsection{Run-time efficiency}
In this section, we test the run-time efficiency. We collect $3886$ rounds of predictions and record the time consumption of the three parts of the system, namely, network inference, cost map rendering, and non-linear optimization. The experiment is conducted on a desktop computer equipped with an Intel I7-8700K CPU and an NVIDIA GTX 1080-Ti graphics card for network training and inference.
\begin{table}[t]
	\caption{Run-Time Analysis}
	\label{tab:runtime_analysis}
	\resizebox{\columnwidth}{!}{
		\begin{tabular}{ |c|c|c|c|c|c|c|c|c|}
			\hline
			\makecell{$\#$ of \\ Predictions }&
			Time(ms)&
			\makecell{Network\\Inference} &
			\makecell{Cost Map\\Rendering} &
			\makecell{Non-linear\\Optimization} &
			\textbf{Total} \\
			\hline
			3886  & \makecell{Avg\\Std\\Max}  &\makecell{\textbf{2.9}\\2.2\\7.2}    & \makecell{\textbf{16.2}\\3.1\\27.0}  & \makecell{\textbf{3.4}\\6.0\\40.0} & \makecell{\textbf{22.6} \\7.6\\68.5}\\
			\hline
		\end{tabular}
	}
	\vspace{-0.8cm}
\end{table}

As we can see from Tab.~\ref{tab:runtime_analysis}, the network inference (on GPU) consumes $2.9$ ms on average since the network structure is not complex. It takes an average computing time of $16.2$ ms to render a 4-layer $40\times40$ 2-D cost map (CPU implementation). 
The non-linear optimization is efficient, with an average time consumption of $3.4$ ms. In total, our prediction system typically consumes $22.6$ ms to complete one round of prediction, and a large part of that time is consumed in the cost map rendering.

\section{Conclusion and Future work}\label{sec:conclusion}
In this paper, we propose an online two-level vehicle trajectory prediction framework which utilizes a policy anticipation network for high-level policy reasoning and a non-linear optimization process for low-level context reasoning. We highlight the flexibility of the proposed framework, and provide various test cases, including normal operations and abnormal driving behavior, in urban environments.
In the future, we will explore using IRL~\cite{abbeel2004apprenticeship} to acquire the weights from data. Modeling interaction in prediction is another direction we are actively exploring~\cite{ding2019int}.
\clearpage
\bibliography{paper}
\end{document}